%% file: main.tex
\documentclass[10pt,twocolumn,letterpaper]{article}

\usepackage{cvpr}
\usepackage{times}
\usepackage{epsfig}
\usepackage{graphicx}
\usepackage{amsmath}
\usepackage{amssymb}
\usepackage{caption}
\usepackage{xspace}

\usepackage[flushleft]{threeparttable}
\usepackage[super]{nth}
\usepackage{booktabs}
\usepackage[pagebackref=true,breaklinks=true,letterpaper=true,colorlinks,bookmarks=false]{hyperref}

\cvprfinalcopy 

\def\cvprPaperID{7} 

\ifcvprfinal\fi

\include{includes}

\begin{document}
\pagenumbering{gobble}
\title{A Large Dataset of Historical Japanese Documents with Complex Layouts}

\author{Zejiang Shen \quad Kaixuan Zhang \quad Melissa Dell\\
Harvard University\\
{\tt\small \{zejiang\_shen, kaixuanzhang, melissadell\}@fas.harvard.edu}}

\twocolumn[{%
\renewcommand\twocolumn[1][]{#1}%
\maketitle
\begin{center}
    \centering
    \includegraphics[width=\linewidth]{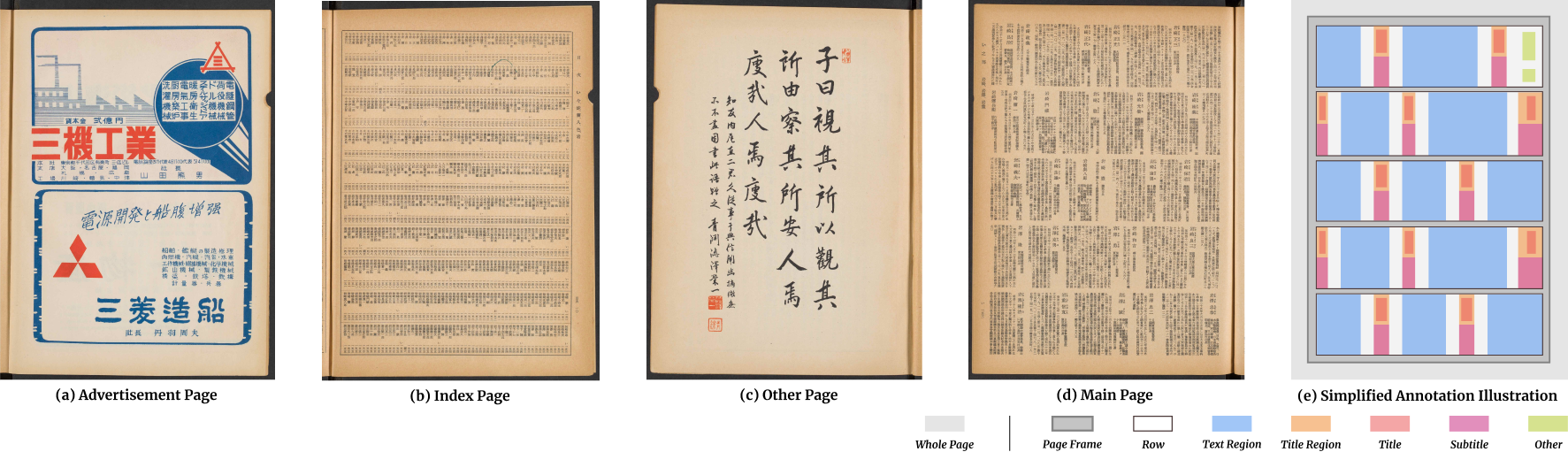}
    \captionof{figure}{\textbf{Examples of \dataname{} document images and annotations}. (a) to (d) show images of the four page categories, and (e) provides a simplified illustration of layout annotations for \imgTypeMain{} pages. The seven types of hierarchically constructed layout elements are highlighted in different colors.}
    \label{fig:overview}
\end{center}%
}]

\begin{abstract}
   \vspace{-1.25mm}
   Deep learning-based approaches for automatic document layout analysis and content extraction have the potential to unlock rich information trapped in historical documents on a large scale. One major hurdle is the lack of large datasets for training robust models. In particular, little training data exist for Asian languages. To this end, we present \dataname, a Large Dataset of \textbf{H}istorical \textbf{J}apanese Documents with Complex Layouts. It contains over 250,000 layout element annotations of seven types. In addition to bounding boxes and masks of the content regions, it also includes the hierarchical structures and reading orders for layout elements. The dataset is constructed using a combination of human and machine efforts. A semi-rule based method is developed to extract the layout elements, and the results are checked by human inspectors. The resulting large-scale dataset is used to provide baseline performance analyses for text region detection using state-of-the-art deep learning models. And we demonstrate the usefulness of the dataset on real-world document digitization tasks. The dataset is available at \dataurl{}.
\end{abstract}


\maketitle

\section{Introduction}

\input{src/1-introduction}

\section{Related Work}

\input{src/2-relatedwork}

\input{src/tables/page_type}
\section{Page Type Labeling}

\input{src/3-pagetype}

\section{Document Layout Annotation}

\input{src/4-method}

\section{Experiments}
\input{src/5-experiments}

\section{Conclusion}

\input{src/6-conclusion}

\paragraph{Acknowledgement.}  This project is supported in part
by NSF Grant \#1823616. 

{\small
\bibliographystyle{ieee_fullname}
\bibliography{ref}
}

\end{document}

%% file: includes.tex
\newcommand{\dataname}{HJDataset}

\newcommand{\dataurl}{\url{https://dell-research-harvard.github.io/HJDataset/}}

\newcommand{\imgTypeMain}{\texttt{main}}
\newcommand{\imgTypeAdv}{\texttt{advertisement}}
\newcommand{\imgTypeIndex}{\texttt{index}}
\newcommand{\imgTypeOther}{\texttt{other}}

\newcommand{\layoutTypePage}{page frame}
\newcommand{\layoutTypeRow}{row}
\newcommand{\layoutTypeTitle}{title region}
\newcommand{\layoutTypeBio}{text region}
\newcommand{\layoutTypeName}{title}
\newcommand{\layoutTypePos}{subtitle}
\newcommand{\layoutTypeOther}{other}

%% file: src/1-introduction.tex
Complex layouts significantly complicate the automated digitization of historical documents, which contain a variety of rich information of interest to researchers and the public more generally. In particular, many documents of relevance to social science researchers and business analysts contain complex, heterogeneous tabular and column structures, which off-the-shelf tools cannot recognize. Moreover, unique layout patterns appear in different languages. For example, complex layouts with vertical text orientation are common in Asian languages.  Complex layouts disrupt Optical Character Recognition (OCR) and result in text from different columns, rows, or text regions being incorrectly garbled together, making automated digitization results unusable.

Various algorithms~\cite{breuel2003high, cattoni1998geometric} have been proposed to analyze the layouts geometrically. They utilize visual properties like text spacing and gaps to correct skewness and segment content regions with fine-tuned parameters. Recently, there has been an increased interest in adopting deep learning (DL) methods to build end-to-end layout understanding models. For example, Oliveira \etal~\cite{oliveira2018dhsegment} and Xu \etal~\cite{ijcai2018-147} build models upon fully convolutional networks~\cite{long2015fully} to detect page frames and text lines with high accuracy.

Central to the success of DL models are many labeled samples for training and evaluating the neural networks. There have been long term efforts to develop layout analysis datasets~\cite{antonacopoulos2009realistic}, and recently a very large-scale dataset has been developed for modern documents~\cite{zhong2019publaynet}. However, for historical documents, the existing datasets are small. For example, there are only 150 instances in the DIVA-HisDB dataset~\cite{simistira2016diva} and 528 in the European Newspapers Project Dataset~\cite{clausner2015enp}. Because deep neural nets tend to overfit small datasets, models trained on them are less robust and performance evaluation is less reliable. Because older documents are subject to wearing, stains, and other noise that do not appear in modern documents, they require dedicated large datasets for training.

Additionally, most open-sourced historical document layout datasets are in western languages~\cite{antonacopoulos2009realistic, clausner2015enp, clausner2019icdar2019}. Models trained on them are not exposed to layout patterns that appear commonly and exclusively in Asian languages. Asian language datasets will be required to build more generalized layout analysis models.

To attack these problems, we present the \dataname: a Large Dataset of \textbf{H}istorical \textbf{J}apanese Documents with Complex Layouts. Currently, this dataset contains 2,271 document image scans with various document information, from the Japanese Who's Who biographical directory published in 1953, which contains biographies for around 50,000 prominent Japanese citizens~\cite{1903Jk}. For each document image, \dataname{} contains its content category (\imgTypeMain{}, \imgTypeIndex{}, \imgTypeAdv{}, or \imgTypeOther{}). For the \imgTypeMain{} and \imgTypeIndex{} pages, we create 25k layout region annotations of seven types at different levels (from page frames to individual text blocks). Besides the bounding box coordinates, we also include the dependency structures and reading orders for all the layout elements.  The data are stored in the COCO format~\cite{lin2014microsoft}, which is commonly used in computer vision research.  The resulting dataset provides a ground truth for different document image analysis tasks, from page classification to layout element detection. Extensive experiments have been conducted, and state-of-the-art models are trained and evaluated on this dataset.

Manual creation of such a dataset would be highly laborious, prohibitively costly, and potentially quite noisy. Therefore, similar to PubLayNet~\cite{zhong2019publaynet}, \dataname{} is generated in a near-automatic fashion. With the help of a carefully designed semi-rule-based method, the layout elements are accurately extracted. To ensure label quality, possible errors are identified based on annotation statistics, and human inspectors correct some minor errors accordingly.

The contribution of this work is twofold. First, we build the \dataname{}, the first large layout analysis dataset of historical Japanese documents to the best of our knowledge. A semi-ruled-based method is designed for generating this dataset. Second, we show that models pre-trained on our dataset can improve performance on other tasks with small amounts of labeled data. The dataset and pre-trained models will be released online to support the development of Japanese and more general layout analysis algorithms.

%% file: src/2-relatedwork.tex
\paragraph{Layout Analysis Dataset} A variety of layout analysis datasets have been created in recent years ~\cite{valveny2014datasets}. For modern documents, Antonacopoulos \etal's work~\cite{antonacopoulos2009realistic} is the first frequently-used dataset, with 305 images of magazines and technical articles available for download. The recent PubLayNet~\cite{zhong2019publaynet} dataset contains 360k samples from modern research publications. For historical documents, the work in \cite{clausner2015enp} provides layout annotations for 600 historical European newspaper images. The datasets in \cite{simistira2016diva} and \cite{gruning2018read} are commonly used for medieval manuscripts and have 160 and 2036 samples respectively. Historical layout datasets tend to be small and are largely unavailable for Asian languages. Large-scale digital libraries, such as the millions of scans placed online by Japan's National Diet Library, provide the raw inputs for creating large datasets for historical document layout analysis, but developing these datasets requires methods that do not rely on costly human labeling.

\paragraph{Deep Learning for Layout Analysis} As deep learning has revolutionized computer vision research, DL-based document image analysis methods are also being developed to tackle challenging tasks. \cite{harley2015evaluation} evaluates convolutional neural networks for document image classification tasks, and \cite{oliveira2018dhsegment} adapts the fully convolutional network (FCN)~\cite{long2015fully} to detect layout element objects inside the page. For more complicated tabular data, Schreiber \etal \cite{schreiber2017deepdesrt} adapt Faster R-CNN~\cite{ren2015faster} and FCN to identify their structures and parse the contents. 
Behind their success, large datasets are required to train and evaluate the models. 

%% file: src/tables/page_type.tex
\begin{table}[h]
\caption{Page types and numbers included in \dataname{}}
\resizebox{1.\linewidth}{!}{
\begin{threeparttable}
        \begin{tabular}{l|cc}
        \toprule
        Page Type     & Number of images & Category ID \tnote{a}\\ \midrule
        \imgTypeMain  & 2048             & 8                     \\
        \imgTypeAdv   & 87               & 9                     \\
        \imgTypeIndex & 82               & 10                    \\
        \imgTypeOther & 54               & 11                    \\ \bottomrule
        \end{tabular}
        \begin{tablenotes}
            \small
            \item[a] As COCO format does not contain an image-level category field, we add a new key for each \textit{image} record called \textit{category id}.
        \end{tablenotes}        
\end{threeparttable}
}
\label{table:pageType}
\vspace{-3mm}
\end{table}

%% file: src/3-pagetype.tex
Contents are organized very differently on pages of different purposes, and hence the first step of the layout analysis pipeline identifies the page type. We manually labeled the page types according to their purposes. As shown in Figure~\ref{fig:overview}, four labels, \ie \imgTypeMain{}, \imgTypeIndex{}, \imgTypeAdv{}, and \imgTypeOther{}, have been created for the 2k images. \imgTypeMain{} pages present the detailed biographical information of around 50,000 Japanese business, political, and cultural leaders with complex structure, forming our primary focus. Table~\ref{table:pageType} provides a detailed description of the classes and the number of samples contained in \dataname{}. 

%% file: src/4-method.tex
\begin{figure}[t]
  \centering
  \includegraphics[width=\linewidth]{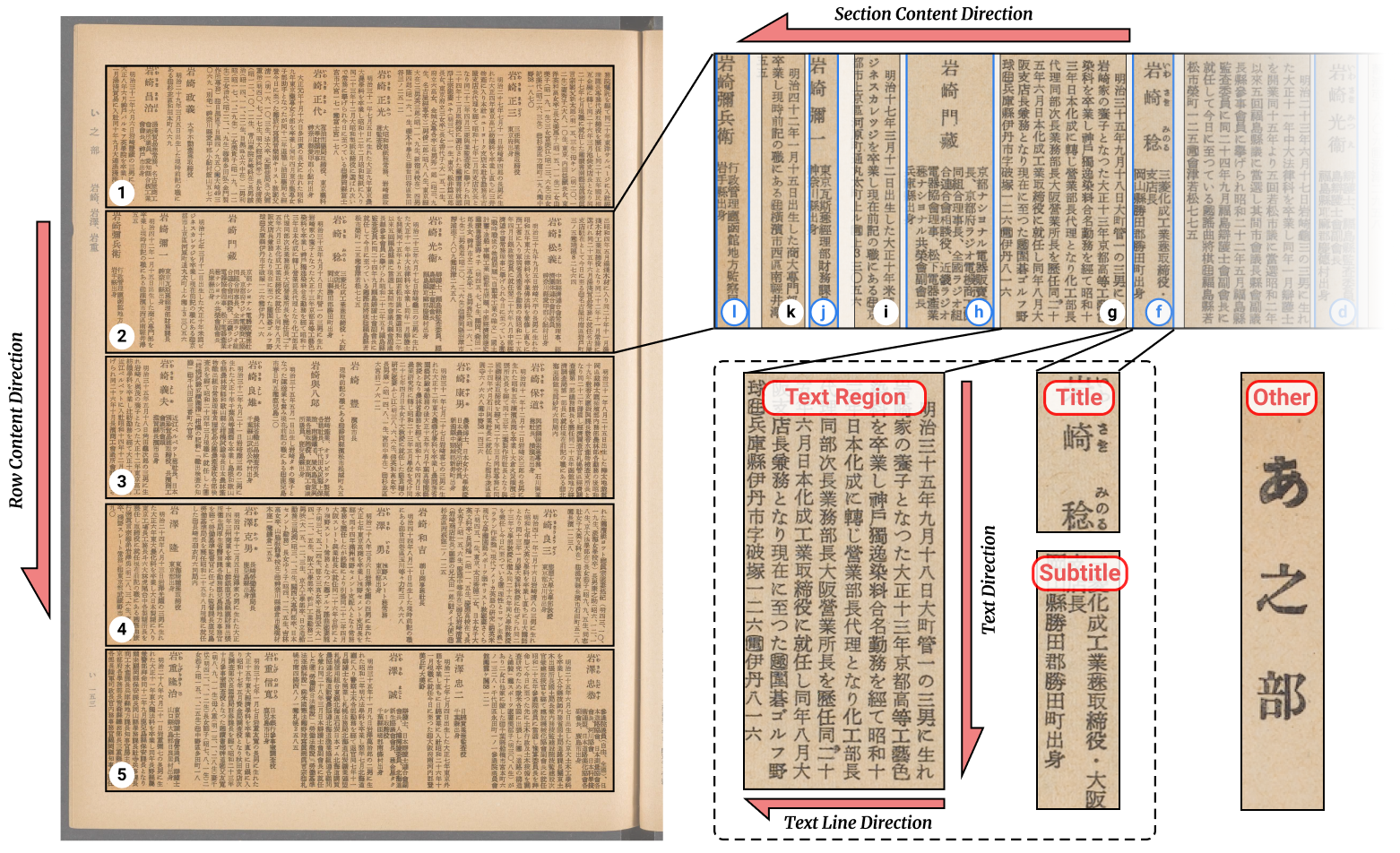}
  \caption{\textbf{The hierarchical content structure in \imgTypeMain{} pages.} Each page contains five rows that are vertically stacked, and the \layoutTypeBio{}s are horizontally arranged within each row. Texts are vertically written inside the \layoutTypeBio{}, \eg (g) in the figure. The \layoutTypeTitle{}, \eg (f), can be further split into \layoutTypeName{} and \layoutTypePos{} blocks. An \layoutTypeOther{} category is reserved for chapter headers and other irrelevant text regions.} 
  \label{fig:layout}
\end{figure}

As shown in Figure~\ref{fig:layout}, the contents in the \imgTypeMain{} pages are organized in a hierarchical manner. Five rows are vertically stacked in a page, while \layoutTypeBio{} and \layoutTypeTitle{} are horizontally arranged inside each row. The \layoutTypeTitle{} can be further broken down into \layoutTypeName{} and \layoutTypePos{} blocks, and other irrelevant texts are labeled as the \textit{other} type. The \layoutTypeBio{} blocks contain only vertical text lines and read from right to left.  Our objective is to segment the pages into units of simple layouts, namely, \layoutTypeBio{}, \layoutTypeName{}, and \layoutTypePos{} blocks. Similar rules apply to the \imgTypeIndex{} pages. 

\begin{figure}[t]
  \centering
  \includegraphics[width=\linewidth]{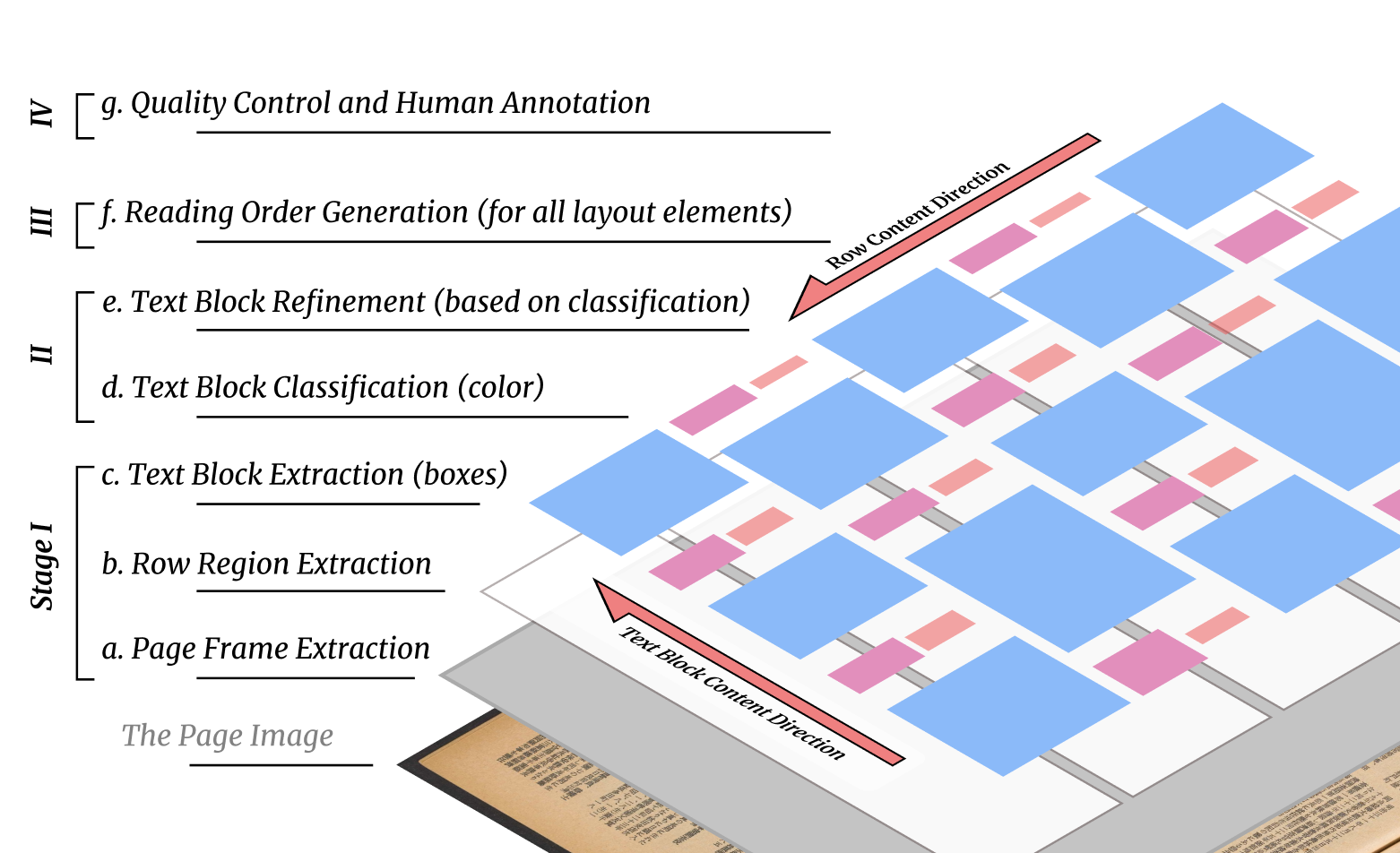}
  \caption{\textbf{The four stages in layout element annotation.} Our method detects the coordinates of the page frames, row regions, and text blocks. A text block classifier is then used to predict the block categories (indicated by the different colors in the figure), and the detections are refined accordingly. Reading orders and hierarchical dependency are generated for all layout elements. Finally human annotators check the results and correct the errors.}
  \label{fig:method}
\end{figure}

Based on the hierarchical structures, we design a multi-stage pipeline for robustly extracting the layout elements, illustrated in Figure~\ref{fig:method}. For the input page scan, the \emph{Text Block Detector} first extracts the bounding boxes of \layoutTypePage{}, \layoutTypeRow{}, \layoutTypeBio{}, and \layoutTypeTitle{} sequentially, as explained in Section~\ref{sec:textBlockDetector}. A CNN is trained to predict the contextual labels for the extracted regions, and the block segmentation is refined accordingly (detailed in Section~\ref{sec:contextualClassifier}). After that, we construct the \emph{reading orders} based on Japanese reading rules, as described in Section~\ref{sec:readingOrder}. Finally, Section~\ref{sec:qualityControl} discusses measures to identify and correct possible errors to ensure high quality of the generated annotations. The dataset statistics are provided in Section~\ref{sec:datasetStats}.

\subsection{Text Block Detector}
\label{sec:textBlockDetector}

The \emph{Text Block Detector} extracts the content boxes in the input scan in a hierarchical fashion. After binarizing the color scans, the recognition is conducted on different resolutions to identify blocks of different scales. The algorithm downsamples the image with a $1/8$ ratio when detecting the \layoutTypePage{} and \layoutTypeRow{} boxes, while using the full resolutions for extracting regions in each row. To account for possible rotations and irregularities, we characterize the \layoutTypePage{} boxes with quadrilaterals. The \layoutTypeRow{}, \layoutTypeBio{}, and \layoutTypeTitle{} are represented with rectangles as the distortions are largely eliminated within the page frames.

As illustrated in Figure~\ref{fig:method}.a, we first estimate the page frame box using contour detection. This method groups pixels with similar visual properties like color or intensity and can be used for extracting different regions~\cite{maire2009contour}.  In our case, the largest intensity contour in the input delineates the page boundary, and we estimate the four vertex coordinates $\left\{(x_i,y_i)\right\}_{i=1}^4$ of the circumscribed quadrilateral for this contour as the page box. We convert the page image inside the quadrilateral to a rectangle based on a warp affine transformation.

Connected Component Labeling (CCL)~\cite{samet1988efficient} and Run Length Smoothing Algorithm (RLSA)~\cite{nikolaou2010segmentation} are used for splitting the five rows of contents vertically inside the page frame. As we apply the RSLA algorithm horizontally, each row is connected, and CCL can be applied to differentiate the rows. This approach is robust when the page is the end of a chapter, where there could be fewer than five rows, and the last row is not ``full''. Similarly, for text and title regions in a row, we apply RLSA vertically and split the connected components. Since the prediction is performed row-wise, it is impossible to connect text blocks in different rows, and the segmentation result is more robust. Rectangular bounding box coordinates $(x_1, y_1, w, h)$ are generated for each row, text and title region, where $(x_1, y_1)$ is the coordinate for the top-left corner, and $w$ and $h$ are width and height for the rectangle, respectively.

\emph{Text Block Detector} finds the layout regions with high accuracy (details in Section~\ref{sec:qualityControl}). However, text and title regions are sometimes mis-segmented due to various noise. Hence, \emph{Text Region Classifier} is developed to identify layout categories and correct segmentation errors.

\begin{figure}[t]
  \centering
  \includegraphics[width=\linewidth]{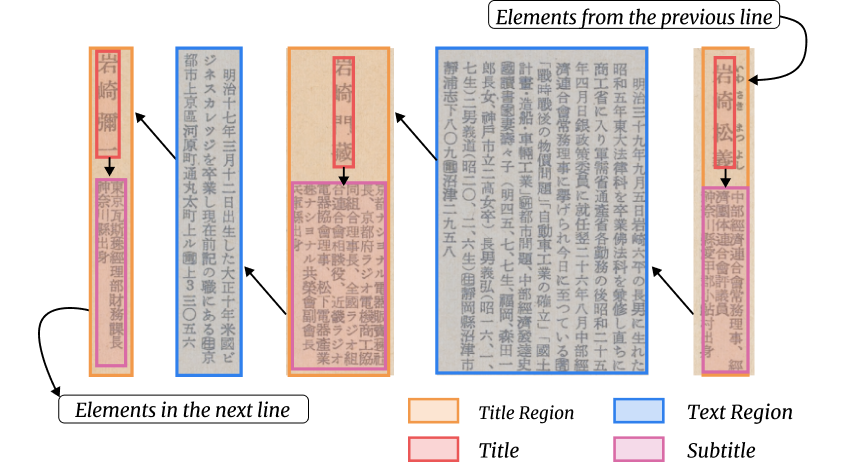}
  \caption{\textbf{Examples of the layout annotations and their reading order.}}
  \label{fig:readingOrder1}
\end{figure}

\subsection{Text Region Classification and Refinement}
\label{sec:contextualClassifier}

A three-class CNN classifier is trained to identify the text, title, and wrongly-segmented regions. After obtaining the region bounding box from \emph{Text Region Detector}, we crop the page image based on the coordinates and predict its category. If it is classified as mis-segmented, a CCL-based method is applied to split it into text and title region. Title regions are further broken down into more refined title and subtitle segments, as illustrated in Figure~\ref{fig:readingOrder1}.

We use the NASNet Mobile~\cite{zoph2018learning} architecture to build our CNN. It is a neural network generated via Neural Architecture Search (NAS) and achieves excellent performance over many benchmarks. Our classifier is trained on 1,200 hand-labeled samples and tested on 100 samples. As mis-segmentation rarely appears (only 3 in 1000 samples), we re-balance the dataset distribution by manually creating 250 mis-segmented images. The input images are rescaled to the same size of 200 height and 522 width. We train the model from scratch, without loading pre-trained weights. Using a stochastic gradient descent optimizer, the loss converges in 40 epochs with a final test accuracy of 0.99.

\begin{figure}[t]
  \centering
  \includegraphics[width=\linewidth]{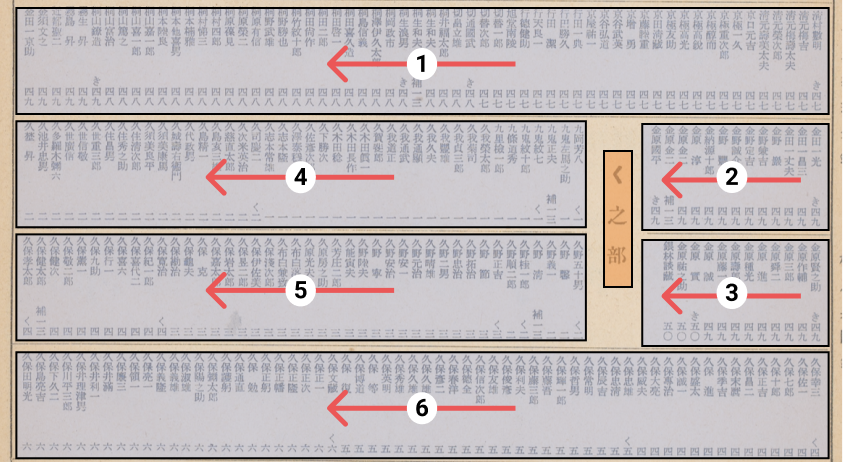}
  \caption{\textbf{Irregular reading orders in the index pages.} The section header in row 2 and 3 disrupts the reading order.}
  \label{fig:readingOrder2}
\end{figure}

\subsection{Reading Order Generation}
\label{sec:readingOrder}

This publication contains non-trivial reading orders which must also be deduced. The texts inside the basic elements (\layoutTypeBio{}, \layoutTypeName{}, and \layoutTypePos{}) are written vertically, and read from right to left. Additionally, for text blocks in a row, they also follow a right-to-left order. Black arrows in Figure~\ref{fig:readingOrder1} shows the topological orders of different elements in a row. 
However, some different structures also exist. As indicated in Figure~\ref{fig:readingOrder2}, section titles (shown in orange) disrupt the regular right-to-left order of texts (see rows 2 and 3). As texts are usually densely arranged in each row, by searching the large gaps between blocks, we identify the discontinuity and correct the special reading order accordingly. We incorporate this irregularity in the dataset to include the real-world noise and support the development of more general layout understanding models.

\subsection{Quality Control and Human Annotations}
\label{sec:qualityControl}

Historical scans are challenging to analyze due to various noise. Despite the carefully engineered method described above, detection errors inevitably exist and need to be handled carefully. However, considering the sheer number of layout elements in this dataset, manual checking of all the predictions would be highly laborious and potentially noisy. 

To identify the small number of incorrect predictions without searching the whole dataset, we examine statistics about blocks and pages. As the \imgTypeMain{} pages are densely printed, we find the number of layout elements remains consistent across pages, and blocks in a row are usually evenly spaced. Hence, by filtering layout elements that are significantly different in these statistics, we obtain a limited number of misdetection candidates. As the specificity (true negative rate) of the subsample is much higher, we can correct the problems more efficiently.  

\begin{figure}[t]
  \centering
  \includegraphics[width=\linewidth]{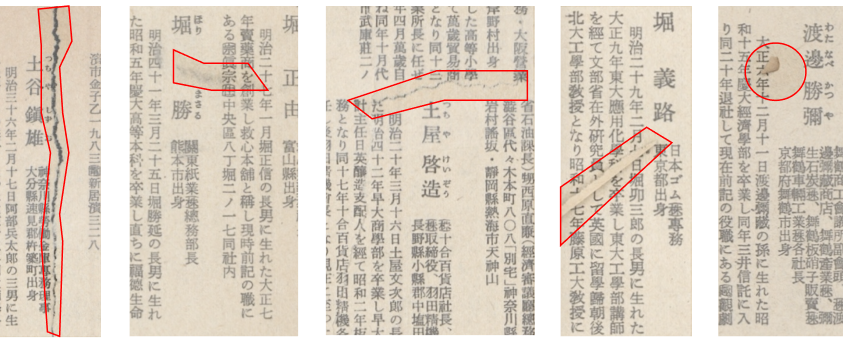}
  \caption{\textbf{Various noises in the page scans.}}
  \label{fig:problems}
  \vspace{4.5mm}
\end{figure}

\paragraph{Misdetected Page Frames} When a page is not appropriately scanned, or it is physically broken, the page frame detection will become inaccurate, and it will disrupt the subsequent extraction of row and text regions. A large increase or decrease in the number of layout elements in a page often implies a misdetection of the page frame. Therefore, we select pages with more than 118 (\nth{95}  percentile) or less than 88 (\nth{5} percentile) layout elements and check them manually. This selects 182 pages, and 18 (9.9\%) errors are identified. Their page frame coordinates are re-labeled manually. After correction, we re-run the pipeline over the pages in order to detect other layout regions more accurately.

\paragraph{Missed Text Lines} The last text lines in a text region are sometimes missed if they contain only a few characters. This results in unusually large gaps between text blocks. This error can be easily identified by filtering the widths of the block gaps. We select 1,011 blocks with gaps larger than 54 pixels (\nth{99} percentile), and correct 487 of them.

\paragraph{Additional Correction} Figure~\ref{fig:problems} shows issues like cracks, stains, and holes that appear frequently and can disrupt the prediction pipeline. It is difficult to pre-screen all the mis-segmentation and incorrect predictions due to these irregularities. Hence, during the manual checking process, human annotators are asked to identify such errors and correct them. A total of 111 layout elements have been found and corrected so far.

In all, we fix more than 616 errors in total (since fixing page frames leads to more improvements), and 80\% are identified by the statistical approach. We estimate that before correction, there are around 1,560 blocks detected inaccurately.\footnote{We randomly choose 20 pages, and count the error rate. This process is repeated 3 times, and the average inaccuracy is 0.6\%, which is equivalent to 1,560 out of 260k blocks.} After correcting the errors, the resultant dataset achieves 99.6\% accuracy, and the remaining 0.4\% errors can be neglected as random noise.

\subsection{Dataset Statistics and Partition}
\label{sec:datasetStats}

A total of 259,616 layout elements of seven categories have been extracted, as detailed in Table~\ref{table:layoutType}. Figure~\ref{fig:data} shows examples of the annotations. Layout elements like \layoutTypeBio{} and \layoutTypeOther{} do not appear in the \imgTypeIndex{} pages, as we characterize the texts in index pages as \layoutTypeName{}.

We partition our dataset into training, validation, and testing subsets: 70\% for training and 15\% each for validation and testing. The breakdown is stratified based on the page type to ensure the equal exposure of different page types in the three subsets. Because the characteristics of the pages vary, categories appear in different frequencies, and the dataset is unbalanced with respect to the object types.

\input{src/tables/layout_stats}

%% file: src/tables/layout_stats.tex
\begin{table}[t]
\caption{Layout element categories and numbers}
\resizebox{1.\linewidth}{!}{
\begin{tabular}{l|cccc}
\toprule
Category & Training & Validation & Test & Total \\ 
\midrule
Page Frame    & 1490  & 320   & 320   & 2130 \\
Row           & 7742  & 1657  & 1660  & 11059 \\
Title Region  & 33637 & 7184  & 7271  & 48092 \\
Text Region   & 38034 & 8129  & 8207  & 54370 \\
Title         & 66515 & 14931 & 14366 & 95812 \\
Subtitle      & 33576 & 7173  & 7256  & 48005 \\
Other         &  103  & 16    & 29    & 148 \\ 
\midrule
Total         & 181097 & 39410 & 39109 & 259616 \\ 
\bottomrule
\end{tabular}
}
\vspace{-5.25mm}
\label{table:layoutType}
\end{table}

%% file: src/5-experiments.tex
In this section, we first report results from training state-of-the-art object detection models on the \dataname{}. Performance is evaluated and provided as a benchmark. Second, based on the pre-trained model, we study how \dataname{} can assist other layout analysis tasks.

\begin{figure*}[t]
  \centering
  \includegraphics[width=\linewidth]{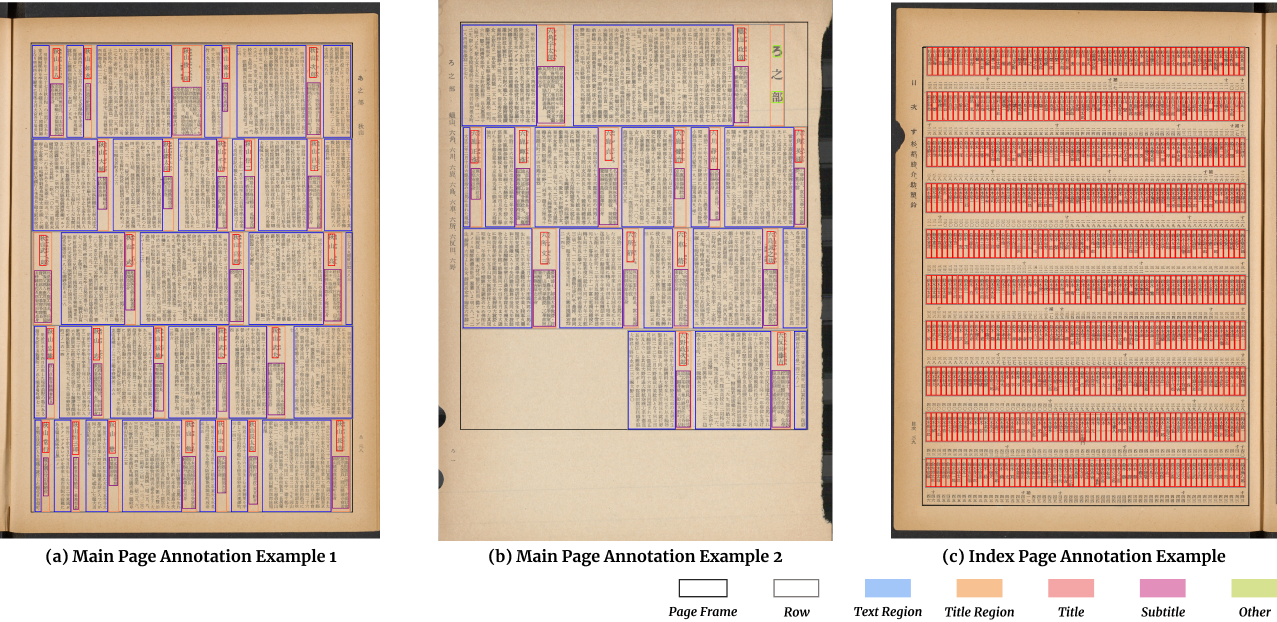}
  \caption{\textbf{Annotation Examples in \dataname{}.} (a) and (b) show two examples for the labeling of \imgTypeMain{} pages. The boxes are colored differently to reflect the layout element categories. Illustrated in (c), the items in each \imgTypeIndex{} page row are categorized as title blocks, and the annotations are denser.}
  \label{fig:data}
  \vspace{-4mm}
\end{figure*}

\subsection{Deep Learning Benchmark}

Without considering the dependency between contents, layout analysis can be treated as \emph{detecting layout objects} inside each page. As object detection has been extensively studied in current deep learning research, well-established models like Faster R-CNN~\cite{ren2015faster}, RetinaNet~\cite{lin2017focal}, and Mask R-CNN~\cite{he2017mask} have achieved excellent performance in various benchmarks~\cite{lin2014microsoft}. Hence, we adopt these models and train them on our dataset. The implementation is based on Detectron2~\cite{wu2019detectron2}, and the neural networks are trained on a single NVIDIA RTX 2080Ti GPU. 

The three models are trained on all layout elements of \imgTypeMain{} pages from the training set. For fair comparison, they are all being trained for 60k iterations, with a base 0.00025 learning rate, and a decay rate of 0.1 for each 30k iterations. The batch size is 2, and the backbone CNN structure is \texttt{R-50-FPN-3x} (details in~\cite{wu2019detectron2}), loaded with pre-trained weights from the COCO dataset. The training configuration will also be open-sourced for reproduciblility.

Table~\ref{table:exp1} shows the per-category bounding box prediction mean Average Precision (mAP) for intersection, at intersection over union (IOU) level [0.50:0.95]\footnote{This is a core metric developed for the COCO competition~\cite{lin2014microsoft} for evaluating the object detection quality.}, on the test data. In general, the high mAP values indicate accurate detection of the layout elements. The Faster R-CNN and Mask R-CNN achieve comparable results, better than RetinaNet. Noticeably, the detections for small blocks like title are less precise, and the accuracy drops sharply for the  title category. In Figure~\ref{fig:exp}, (a) and (b) illustrate the accurate prediction results of the Faster R-CNN model.




\subsection{Pre-training for other datasets}

We also examine how our dataset can help with a real-world document digitization application. When digitizing new publications, researchers usually do not generate large scale ground truth data to train their layout analysis models. If they are able to adapt our dataset, or models trained on our dataset, to develop models on their data, they can build their pipelines more efficiently and develop more accurate models. To this end, we conduct two experiments. First we examine how layout analysis models trained on the \imgTypeMain{} pages can be used for understanding \imgTypeIndex{} pages. Moreover, we study how the pre-trained models perform on other historical Japanese documents. 
\input{src/tables/main_exp}

\begin{figure*}[t]
  \centering
  \includegraphics[width=\linewidth]{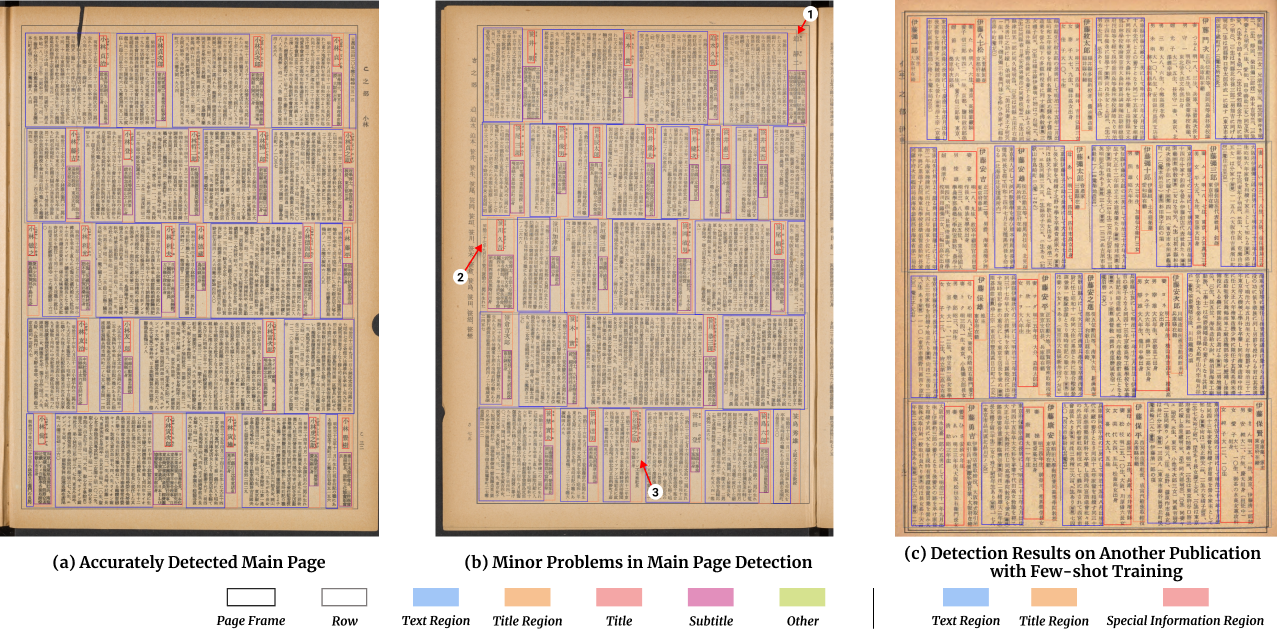}
  \caption{\textbf{The prediction results of Faster R-CNN on Main pages in \dataname{} and another publication.} (a) shows that the Faster R-CNN model is robust to noise like cracks and can detect most of the layout elements accurately. (b) highlights some minor errors in the Faster R-CNN predictions like inaccurate row blocks, \eg (1), and missed text and title regions, \eg (2) and (3). (c) shows the results of few-shot trained Faster R-CNN on another publication. They are generally correct. We label the new publication differently to increase the difficulty for training, and the red boxes in the image denote a special information region.}
  \label{fig:exp}
  \vspace{-3mm}
\end{figure*}

Table~\ref{table:exp2} compares the performance of five Faster R-CNN models that are trained differently on \imgTypeIndex{} pages. If the model loads pre-trained weights from HJDataset, it includes information learned from \imgTypeMain{} pages. Models trained over all the training data can be viewed as the benchmarks, while training with few samples (five in this case) are considered to mimic real-world scenarios. Given different training data, models pre-trained on \dataname{} perform significantly better than those initialized with COCO weights. Intuitively, models trained on more data perform better than those with fewer samples. We also directly use the model trained on \imgTypeMain{} to predict \imgTypeIndex{} pages without fine-tuning. The low zero-shot prediction accuracy indicates the dissimilarity between \imgTypeIndex{} and \imgTypeMain{} pages. The large increase in mAP from 0.344 to 0.471 after the model is trained on five samples shows that the model can be quickly adapted to similar tasks. As the AP\textsubscript{50} and AP\textsubscript{75} (AP calculated with IOU=0.50 and 0.75) are higher than mAP, we conclude that the models can learn to detect the \textit{general} position of layout objects. 

\input{src/tables/index_exp}

To evaluate our models on other historical Japanese documents, we manually annotate 12 pages from another publication with different layouts, the Japanese Who’s Who biographical directory published in 1939~\cite{1939Jk}, and we train the models on 4 samples. Performance is assessed on the remaining 8 samples, as reported in Table~\ref{table:exp3}. Similar to the previous experiment, pre-training on \dataname{} has a large positive influence on the detection accuracy given few training samples. And shown in Figure~\ref{fig:exp} (c), the layout elements are detected accurately. In summary, these two experiments demonstrate the usefulness of our dataset for other layout analysis tasks.

\input{src/tables/other_exp}

%% file: src/tables/main_exp.tex
\begin{table}[b]
\caption{Detection mAP @ IOU [0.50:0.95] of different models for each category on the test set. All values are given as percentages.}
\resizebox{1.\linewidth}{!}{
\begin{threeparttable}
        \begin{tabular}{l|ccc}
            \toprule
            Category & Faster R-CNN & Mask R-CNN\tnote{a} & RetinaNet \\ 
            \midrule
            Page Frame    & 99.046 & 99.097 & 99.038 \\
            Row           & 98.831 & 98.482 & 95.067 \\
            Title Region  & 87.571 & 89.483 & 69.593 \\
            Text Region   & 94.463 & 86.798 & 89.531 \\
            Title         & 65.908 & 71.517 & 72.566 \\
            Subtitle      & 84.093 & 84.174 & 85.865 \\
            Other         & 44.023 & 39.849 & 14.371 \\ 
            \midrule
            mAP           & 81.991 & 81.343 & 75.223 \\
            \bottomrule
            \end{tabular}
        \begin{tablenotes}
            \small
            \item[a] For training Mask R-CNN, the segmentation masks are the quadrilateral regions for each block. Compared to the rectangular bounding boxes, they delineate the text region more accurately.
        \end{tablenotes}        
\end{threeparttable}
}
\label{table:exp1}
\end{table}

%% file: src/tables/index_exp.tex
\begin{table}[t]
\caption{Comparison of the test set AP of Faster R-CNN models trained differently on \imgTypeIndex{} pages. All values are given as percentages.}

\resizebox{1.\linewidth}{!}{
\begin{threeparttable}

    \begin{tabular}{l|c|ccc}
    \toprule
    Initialization & Training Data & mAP & AP\textsubscript{50}    & AP\textsubscript{75}\\
    \midrule
    COCO          & All\tnote{a}  & 34.408 & 53.342 & 37.533 \\
    COCO          & Few-shot      & 9.988  & 18.572 &  9.669 \\
    HJDataset     & All           & 47.125 & 67.502 & 54.410 \\
    HJDataset     & Few-shot      & 10.275 & 21.353 & 10.423 \\
    HJDataset     & Zero-shot     &  9.411 & 44.299 & 0.068  \\
    \bottomrule
    \end{tabular}
    
    \begin{tablenotes}
        \small
        \item[a] \textit{All} indicates the model is trained on all 57 training \imgTypeIndex{} samples, \textit{few-shot} refers to model trained on 5 random samples, and \textit{zero-shot} means the model directly use the weights without training. 
    \end{tablenotes} 
    
\end{threeparttable}
}
\vspace{-2mm}
\label{table:exp2}
\end{table}

%% file: src/tables/other_exp.tex
\begin{table}[t]
\caption{Comparison of the test set AP of Faster R-CNN models trained differently on another publication. All values are given as percentages.}
\resizebox{1.\linewidth}{!}{
\begin{threeparttable}
    \begin{tabular}{l|c|ccc}
    \toprule
    Initialization & Training Data & mAP & AP\textsubscript{50}    & AP\textsubscript{75}\\
    \midrule
    COCO          & Few-shot      &  69.925 & 95.119 & 78.667   \\
    HJDataset     & Few-shot      &  81.638 & 98.364 & 88.203   \\
    HJDataset     & Zero-shot     &  38.959 & 50.971 & 42.269    \\
    \bottomrule
    \end{tabular}
\end{threeparttable}
}
\vspace{-2mm}
\label{table:exp3}
\end{table}


%% file: src/6-conclusion.tex
In this paper, we introduce the HJDataset, a large layout analysis dataset for historical Japanese documents. With a combination of semi-rule-based segmentation and statistical error identification and correction, 260k layout annotations of seven categories are extracted from 2.2k page scans. Page type labels, block dependency, and reading orders are also included. Stored in COCO format, \dataname{} allows state-of-the-art object detection models to be easily trained and evaluated. Moreover, we show that deep learning models trained on \dataname{} can be adapted to other datasets, facilitating real-world document digitization tasks. 